\documentclass[10pt,twocolumn,letterpaper]{article}

\usepackage{cvpr}              

\usepackage{graphicx}
\usepackage{amsmath}
\usepackage{amssymb}
\usepackage{booktabs}
\usepackage{siunitx}
\usepackage{cite}
\usepackage{ctable}
\usepackage{hhline}
\usepackage{booktabs} 
\usepackage{subcaption}
\usepackage{csquotes}
\usepackage{multirow}
\usepackage{pifont}
\usepackage[super]{nth}
\usepackage{paralist}

\newcolumntype{L}[1]{>{\raggedright\let\newline\\\arraybackslash\hspace{0pt}}m{#1}}
\newcolumntype{C}[1]{>{\centering\let\newline\\\arraybackslash\hspace{0pt}}m{#1}}
\newcolumntype{R}[1]{>{\raggedleft\let\newline\\\arraybackslash\hspace{0pt}}m{#1}}

\usepackage[pagebackref,breaklinks,colorlinks]{hyperref}
\usepackage[capitalize]{cleveref}
\crefname{section}{Sec.}{Secs.}
\Crefname{section}{Section}{Sections}
\Crefname{table}{Table}{Tables}
\crefname{table}{Tab.}{Tabs.}

\def\confName{CVPR}
\def\confYear{2022}

\begin{document}
\pagestyle{plain}

\title{Accurate 3D Hand Pose Estimation for Whole-Body 3D Human Mesh Estimation}

\author{
Gyeongsik Moon$^{1}$\hspace{1.0cm} Hongsuk Choi$^{1}$\hspace{1.0cm} Kyoung Mu Lee$^{1,2}$\\
\\
$^{1}$Dept. of ECE \& ASRI, $^{2}$IPAI, Seoul National University, Korea \hspace{1.0cm}
\\
{\small \texttt {\{mks0601,redarknight,kyoungmu\}@snu.ac.kr}}
}

\maketitle

\begin{abstract}
Whole-body 3D human mesh estimation aims to reconstruct the 3D human body, hands, and face simultaneously.
Although several methods have been proposed, accurate prediction of 3D hands, which consist of 3D wrist and fingers, still remains challenging due to two reasons.
First, the human kinematic chain has not been carefully considered when predicting the 3D wrists.
Second, previous works utilize body features for the 3D fingers, where the body feature barely contains finger information.
To resolve the limitations, we present Hand4Whole, which has two strong points over previous works.
First, we design Pose2Pose, a module that utilizes joint features for 3D joint rotations.
Using Pose2Pose, Hand4Whole utilizes hand MCP joint features to predict 3D wrists as MCP joints largely contribute to 3D wrist rotations in the human kinematic chain.
Second, Hand4Whole discards the body feature when predicting 3D finger rotations.
Our Hand4Whole is trained in an end-to-end manner and produces much better 3D hand results than previous whole-body 3D human mesh estimation methods.
The codes are available here
\footnote{\url{https://github.com/mks0601/Hand4Whole_RELEASE}}\textsuperscript{,}\footnote{\url{https://github.com/mks0601/Pose2Pose_RELEASE}}.
\end{abstract}

\section{Introduction}

Whole-body 3D human mesh estimation aims to localize mesh vertices of all human parts, including body, hands, and face, simultaneously in the 3D space.
By combining 3D mesh of all human parts, we can understand not only human body pose and shape but also human intention and feeling through hand poses and facial expressions.
The challenges of this problem are small image sizes and complicated articulations of hands, and smoothly connecting estimated 3D body and hands meshes.

\begin{figure}[t]
\begin{center}
\includegraphics[width=0.7\linewidth]{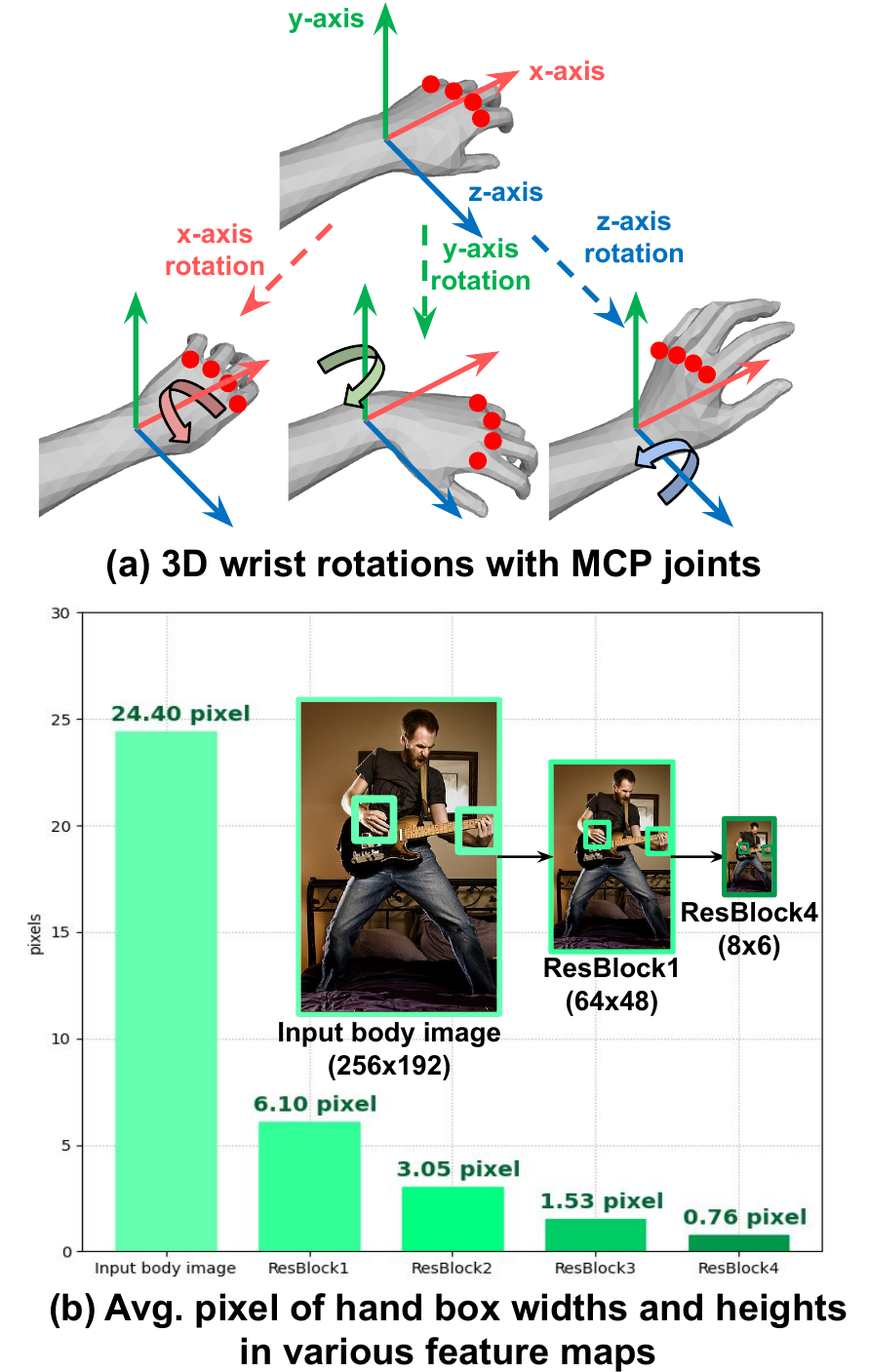}
\end{center}
\vspace*{-5mm}
\caption{
(a) Four hand MCP joints (red circles) provide essential 3D wrist rotation information as they are child nodes of a wrist in hand kinematic chain.
(b) The average of width and height (pixel) of hand boxes in MSCOCO~\cite{jin2020whole}.
The detailed finger information is almost missing at the output of the standard backbone, ResNet (less than 1 pixel).
}
\vspace*{-3mm}
\label{fig:motivation}
\end{figure}

\begin{figure*}[t]
\begin{center}
\includegraphics[width=\linewidth]{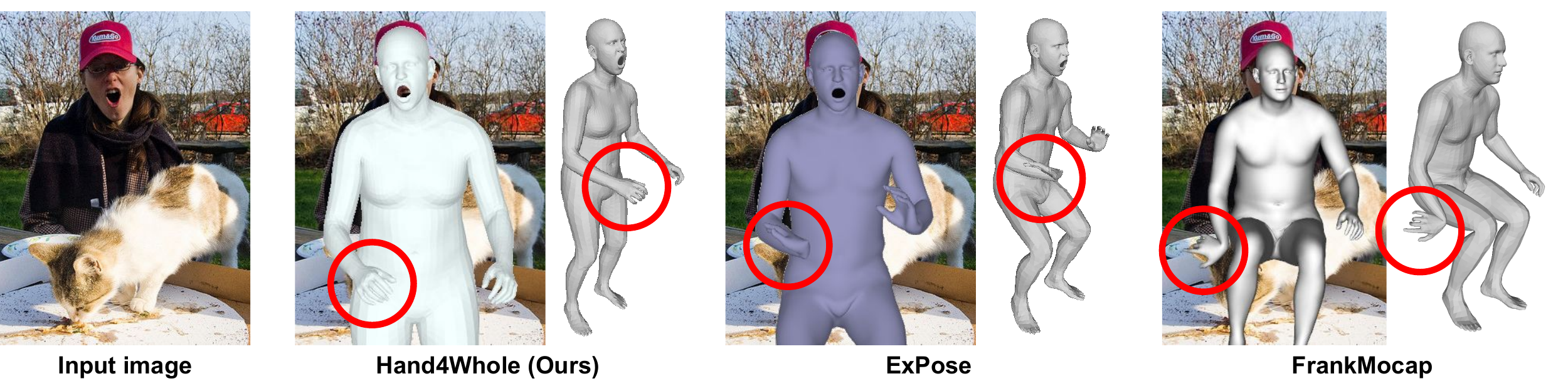}
\end{center}
\vspace*{-5mm}
\caption{
Qualitative results comparison between the proposed Hand4Whole, ExPose~\cite{choutas2020monocular}, and FrankMocap~\cite{rong2021frankmocap} when hands are invisible.
Taking a body feature is necessary for plausible 3D wrist rotations.
}
\vspace*{-3mm}
\label{fig:intro_compare}
\end{figure*}

Existing whole-body 3D human mesh estimation systems~\cite{zhou2021monocular,choutas2020monocular,rong2021frankmocap,feng2021collaborative} consist of body, hands, and face networks.
As hands and face take small areas in the human image, the previous works crop and resize the hands and face images to higher resolutions.
Then, they process the human, hands, and face images using body, hand, and face networks, respectively.
Their body, hand, and face networks perform global average pooling (GAP) to the extracted image features and predict 3D body, hands, and face, respectively, which consist of 3D joint rotations and other parameters (\textit{e.g.}, body shape and facial expressions).
The 3D body, hands, and face are passed to SMPL-X layer~\cite{pavlakos2019expressive} for the final whole-body 3D mesh.

3D hands recovery consists of the 3D wrist and finger rotation predictions.
Accurate 3D hands recovery is a key for the whole-body 3D human mesh estimation; however, previous works~\cite{zhou2021monocular,choutas2020monocular,rong2021frankmocap,feng2021collaborative} produce inaccurate 3D hands due to two reasons.
First, for the 3D wrist rotation, they have not carefully considered the human kinematic chain.
In the human kinematic chain, the wrist connects the human body and fingers' root joints (\textit{i.e.}, hand MCP joints); therefore, utilizing human body and MCP joint information is necessary for accurate 3D wrist rotation.
In particular, the MCP joint information provides the necessary cue to determine 3D wrist rotations, as shown in Figure~\ref{fig:motivation} (a).
Furthermore, the body information provides overall body posture, which can make the predicted 3D wrist rotation be anatomically plausible and smoothly connected with the body, even when hands are occluded or truncated, as shown in Figure~\ref{fig:intro_compare}.
However, none of the previous works used a combination of the body feature and MCP joint features for the 3D wrist rotations.
All of them used one of the body feature~\cite{zhou2021monocular}, hand feature~\cite{choutas2020monocular,rong2021frankmocap}, and their combination~\cite{feng2021collaborative}, where the hand feature contains lots of unnecessary information, such as finger information, while the MCP joint feature contains essential 3D wrist rotation information.
In particular, the finger information can hurt the 3D wrist rotation accuracy as the fingers often move independently to the wrists and have highly complicated articulations.

Second, for the 3D finger rotations, previous works~\cite{zhou2021monocular,feng2021collaborative} used body features in addition to hand features.
The body feature mainly contains much unnecessary information, such as body and backgrounds, while having very coarse hand information due to the small sizes of hands, as shown in Figure~\ref{fig:motivation} (b).
Therefore, such unnecessary information can corrupt the 3D finger rotation accuracy.

To resolve the above limitations, we present Hand4Whole, a whole-body 3D human mesh estimation system that produces much better 3D hands in the whole-body 3D mesh.
Hand4Whole has two stronger points than previous works.
First, we design Pose2Pose, a 3D positional pose-guided 3D rotational pose prediction framework.
Here, the 3D positional and rotational pose represent 3D joint positions and rotations, respectively.
In contrast to previous works~\cite{choutas2020monocular,rong2021frankmocap,feng2021collaborative} that vectorize the image feature by performing GAP, ours extract joint features by our positional pose-guided pooling (PPP).
The joint features enable Pose2Pose to understand different semantic information of each joint, while the previously used vectorized image feature only provides only an instance-level understanding.
Using Pose2Pose, Hand4Whole utilizes a combination of the body feature and eight MCP joint features of two hands for the 3D wrist rotations.
The body feature makes the 3D wrist rotations anatomically plausible when hands are occluded or truncated, as shown in Figure~\ref{fig:intro_compare}.
Also, from an anatomical point of view, the eight hand MCP joint features provide essential 3D wrist rotations among thirty finger joints of two hands, as shown in Figure~\ref{fig:motivation} (a).
By combining both features, Hand4Whole produces much more accurate 3D wrist rotations.

Second, we discard body features when predicting 3D finger rotations.
As a result, we can prevent the coarse hand information from corrupting the 3D finger rotation accuracy.
Although we do not use the body feature for 3D finger rotation, the 3D hands can be anatomically plausible and smoothly connected with body joints as our 3D wrist rotations already consider the body network's feature.
In particular, the wrists are root nodes of a human hand kinematic chain, which determine global 3D rotations of fingers. 
Our Hand4Whole is trained in an end-to-end manner and significantly outperforms previous whole-body 3D human mesh estimation methods.

Our contributions can be summarized as follows.
\begin{itemize}
\item Our Hand4Whole produces much more accurate 3D hands, which consists of the 3D wrist and finger rotations.
Using our Pose2Pose, Hand4Whole utilizes both body and hand MCP joint features for accurate 3D wrist rotation and smooth connection between 3D body and hands.
\item In addition, we discard body features when predicting 3D finger rotations.
\item Hand4Whole is trained in an end-to-end manner and largely outperforms previous whole-body 3D human mesh estimation methods.
\end{itemize}
\section{Related works}

\begin{figure*}[t]
\begin{center}
\includegraphics[width=\linewidth]{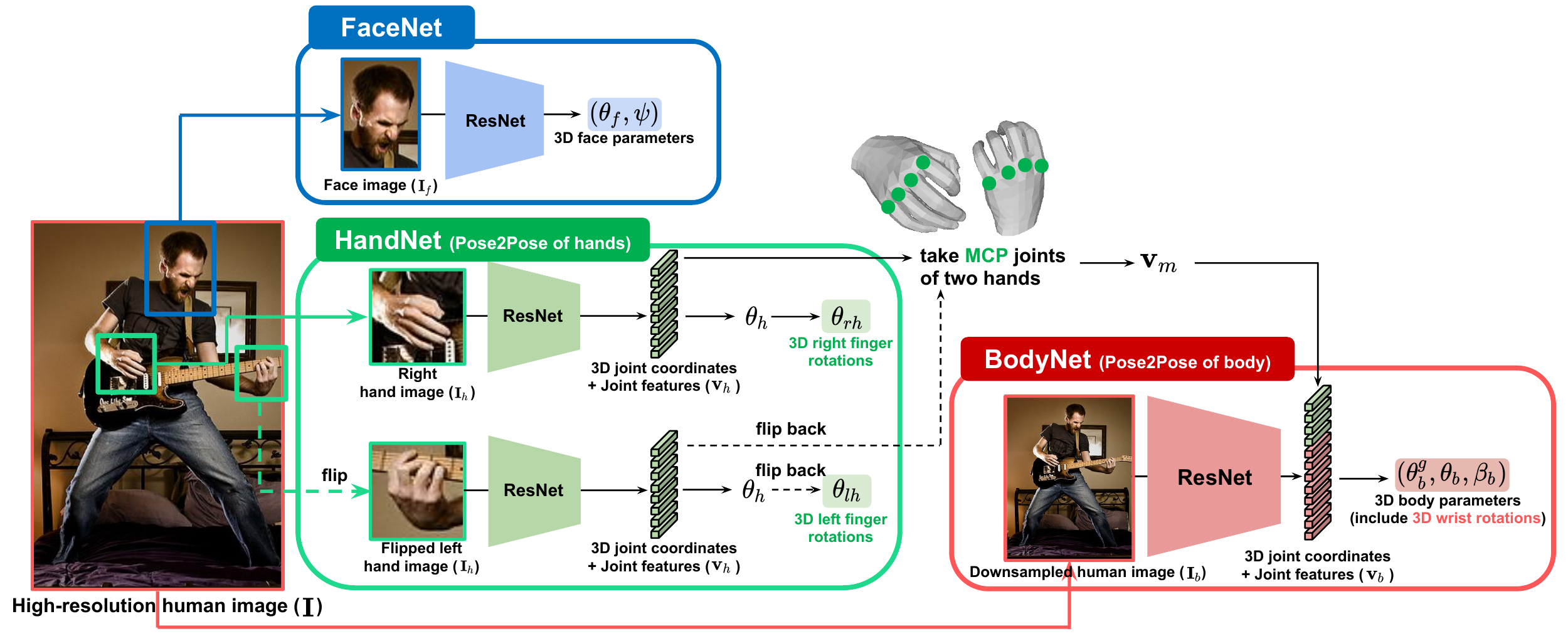}
\end{center}
\vspace*{-5mm}
   \caption{
   The overall pipeline of Hand4Whole.
   The BodyNet predicts 3D body parameters, which include 3D wrist rotations, from a combination of body and hand MCP joint information.
   The HandNet predicts 3D finger rotation only from fine hand information without combining coarse hand information, extracted from the BodyNet.
   The weights in HandNet for the right and flipped left hand images are shared.
   The final whole-body 3D human mesh is obtained by forwarding the outputs of BodyNet, HandNet, and FaceNet to the SMPL-X~\cite{pavlakos2019expressive} layer.
   The hand and face boxes are predicted from BodyNet, not shown in the figure.
   Our Hand4Whole is trained in an end-to-end manner.}
\label{fig:overall_pipeline}
\vspace*{-5mm}
\end{figure*}

Although there have been remarkable progress in 3D body-only~\cite{kanazawa2018end,kolotouros2019learning,moon2020i2l,choi2020p2m,muller2021self,ROMP,lin2021end}, hand-only~\cite{boukhayma20193d,ge20193d,kulon2020weakly,zhou2020monocular}, and face-only~\cite{sanyal2019learning} mesh estimation, there have been few attempts to simultaneously recover the 3D body, hands, and face.
Several early works are based on an optimization-based approach, which fits a 3D human model to the 2D/3D evidence.
Joo~\etal~\cite{joo2018total} fits their human models (\textit{i.e.}, Frank and Adam) to 3D human joints coordinates and point clouds in a multi-view studio environment.
Xiang~\etal~\cite{xiang2019monocular} extended Joo~\etal~\cite{joo2018total} to the single RGB case.
SMPLif-X~\cite{pavlakos2019expressive} and Xu~\etal~\cite{xu2020ghum} fit their human models, SMPL-X and GHUM, respectively, to 2D human joint coordinates.
As the above optimization-based methods can be slow and prone to noisy evidence, a regression-based approach is presented recently.
Recently, several neural network-based methods have been proposed.
All of them consist of body, hand, and face networks, where each takes a human image, hand-cropped image, and face-cropped image, respectively.
The three networks predict SMPL-X parameters for a final whole-body 3D human mesh.
ExPose~\cite{choutas2020monocular} is the earliest regression-based approach, which utilizes the three separated networks.
FrankMocap~\cite{rong2021frankmocap} consists of body and hand networks, and their hand prediction is attached to 3D body results.
Zhou~\etal~\cite{zhou2021monocular} utilize 3D joint coordinates for the 3D joint rotation prediction.
PIXIE~\cite{feng2021collaborative} introduced a moderator to predict 3D joint rotations.

Compared to the recent whole-body 3D human mesh estimation methods, our Hand4Whole has three clear novel contributions.
First, we design Pose2Pose as the main module of our Hand4Whole.
Pose2Pose enables Hand4Whole to understand different semantic information of each joint by introducing joint features, while previous works~\cite{choutas2020monocular,rong2021frankmocap,feng2021collaborative} provides only an instance-level feature to their system by performing the GAP.
Second, Hand4Whole uses both body and hand MCP joint features for the 3D wrist rotations.
On the other hand, ExPose~\cite{choutas2020monocular} and FrankMocap~\cite{rong2021frankmocap} predict 3D wrist rotation only from a hand feature without a body feature.
The absence of the body feature when predicting the 3D wrist rotations results in implausible 3D wrist rotations when hands are occluded or truncated, as shown in Figure~\ref{fig:intro_compare}.
Zhou~\etal~\cite{zhou2021monocular} predict 3D wrist rotations only from a body feature without a hand feature.
They suffer from inaccurate 3D wrist rotations as they do not utilize MCP joint features.
PIXIE~\cite{feng2021collaborative} combines body and hand features by a moderator to predict 3D wrist rotations.
However, their hand feature contains lots of unnecessary information, such as finger information, which can hurt the 3D wrist rotation accuracy as fingers often move independently to the wrists.
Third, Hand4Whole discards the body feature when predicting 3D finger rotations.
In contrast, Zhou~\etal~\cite{zhou2021monocular} and PIXIE~\cite{feng2021collaborative} uses the body feature, which contains very coarse hand information due to the small pixel size of the hands.

\section{Hand4Whole}

Figure~\ref{fig:overall_pipeline} shows the overall pipeline of the proposed Hand4Whole for whole-body 3D human mesh estimation.
It consists of BodyNet, HandNet, and FaceNet, which take cropped and resized human, hands, and face images, respectively.
The outputs of each network are fed to SMPL-X~\cite{pavlakos2019expressive} layer to obtain the final whole-body 3D human mesh.

\subsection{Pose2Pose}~\label{sec:pose2pose}
\begin{figure}[t]
\begin{center}
\includegraphics[width=\linewidth]{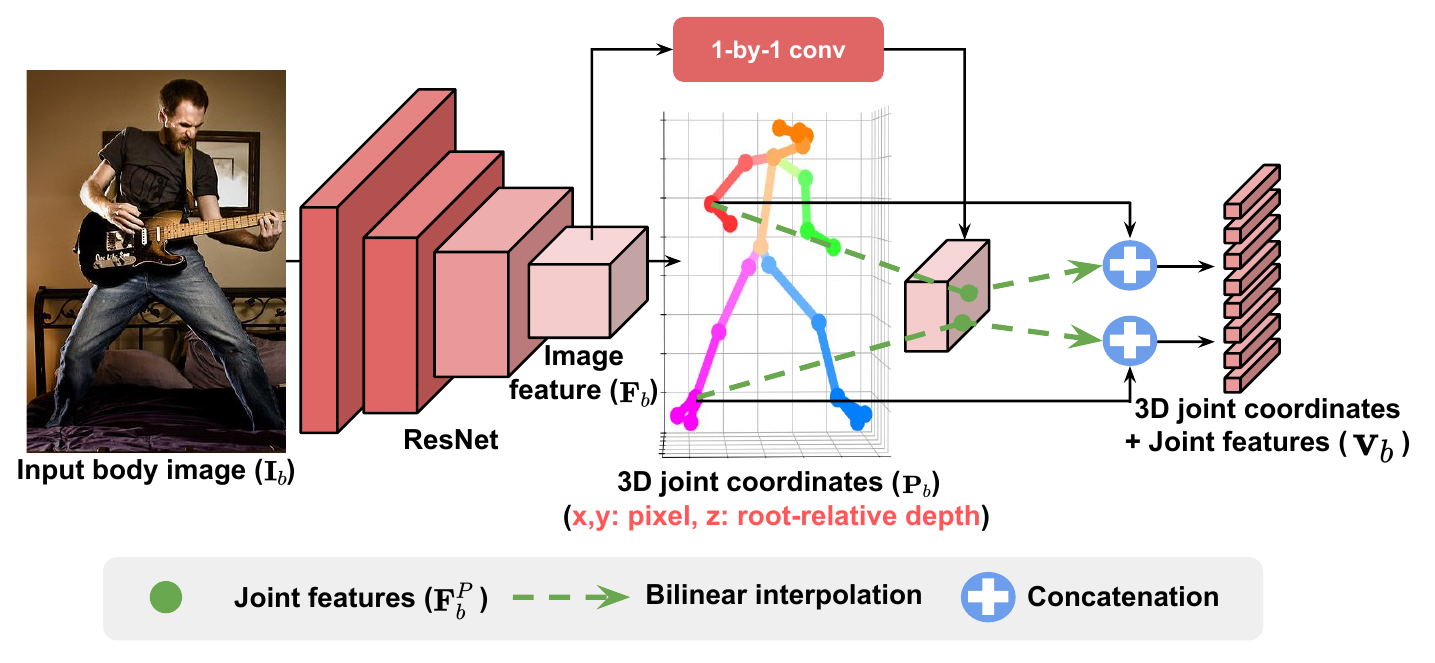}
\end{center}
\vspace*{-5mm}
   \caption{
   Illustration of the Pose2Pose in our BodyNet.
   It extracts the joint features by interpolating $(x,y)$ pixel position of the 3D joint coordinates $\mathbf{P}_b$ on the output of 1-by-1 convolution.
   For the simplicity, we describe only right elbow and ankle.}
\label{fig:bodynet}
\vspace*{-3mm}
\end{figure}

To start, we describe Pose2Pose, a 3D positional pose (\textit{i.e.}, 3D joint positions)-guided 3D rotational pose (\textit{i.e.}, 3D joint rotations) prediction framework.
Figure~\ref{fig:bodynet} shows the overall pipeline of Pose2Pose when it is used for the 3D body.
Pose2Pose enables the proposed Hand4Whole to understand different semantic information of each joint.
We use Pose2Pose as the main module of our BodyNet and HandNet.
Pose2Pose consists of two stages, described in the following.

\noindent\textbf{3D joint coordinates estimation.}
Pose2Pose firsts predicts the 3D joint coordinates $\mathbf{P} \in \mathbb{R}^{J \times 3}$ from a human image.
$J$ denotes the number of joints.
$x$- and $y$-axis of $\mathbf{P}$ are in pixel space, and $z$-axis of it is in root joint (\textit{i.e.}, pelvis for the body and wrist for the hand)-relative depth space.
To this end, ResNet-50~\cite{he2016deep} extracts 2048-dimensional image feature map $\mathbf{F}$ from the input image.
We use ResNet after removing the GAP and fully-connected layer of the last part of the original ResNet.
Then, a 1-by-1 convolutional layer predicts 3D heatmaps of human joints $\mathbf{H}$ from $\mathbf{F}$, where $\mathbf{H}$ has the same height and width as $\mathbf{F}$.
To make 3D heatmaps from the 2D feature map, the 1-by-1 convolutional layer changes the channel dimension from 2048 to $8J$.
Then, we reshape the $8J$-dimensional feature map to $J$ dimensional 3D heatmap whose depth size is 8.
The 3D joint coordinates $\mathbf{P}$ is obtained from $\mathbf{H}$ by the soft-argmax operation~\cite{sun2018integral} in a differentiable way.

\noindent\textbf{Positional pose-guided pooling (PPP).}
Positional pose-guided pooling extracts joint features from the 2D feature map $\mathbf{F}$.
To this end, we apply a 1-by-1 convolution to $\mathbf{F}$ to change its channel dimension from 2048 to 512.
Then, we perform a bilinear interpolation to the output of the convolution at $(x,y)$ position of 3D joint coordinates $\mathbf{P}$, which produce joint features $\mathbf{F}^P \in \mathbb{R}^{J \times 512}$.
Please note that as $(x,y)$ coordinates of $\mathbf{P}$ are in pixel space, we use them directly for the interpolation without a 3D-to-2D projection.
The joint features provide different contextual information of each joint, useful for the 3D joint rotation~\cite{guler2019holopose}.

\noindent\textbf{3D joint rotations estimation.}
For the 3D joint rotation $\theta$, we concatenate the joint features $\mathbf{F}^P$ and 3D joint coordinates $\mathbf{P}$ and flatten it, denoted by $\mathbf{v} \in \mathbb{R}^{512J+3J}$.
Then, $\mathbf{v}$ is passed to a fully-connected layer, which produces the 3D joint rotations $\theta$.

\subsection{BodyNet}~\label{sec:bodynet}
The BodyNet outputs 3D body joint rotations, SMPL-X shape parameters, camera parameters, and boxes of hands and face.
Pose2Pose plays a central role in the prediction of the 3D body joint rotation.

\noindent\textbf{3D body joint rotations.}
Using the Pose2Pose, BodyNet predicts 3D body joint rotation $\theta_b \in \mathbb{R}^{22 \times 3}$ from a human image $\mathbf{I}_b \in \mathbb{R}^{3 \times 256 \times 192}$, downsampled from a high-resolution human image $\mathbf{I} \in \mathbb{R}^{3 \times 512 \times 384}$.
The downsampling is necessary to save the computational cost.
One modification we made to the Pose2Pose when predicting the 3D joint rotation is that instead of only using $\mathbf{v}_b$, a concatenation of 3D body joint coordinates and body joint features, BodyNet additionally uses $\mathbf{v}_m$, a concatenation of 3D coordinates and features of MCP joints.
$\mathbf{v}_m$ is obtained from the HandNet, and Section~\ref{sec:handnet} describes how we obtain it in detail.
The additional usage of MCP joint information enables BodyNet to produce more accurate 3D wrist rotations.
Please note that the predicted 3D body joint rotations $\theta_b$ include 3D wrist rotations.
The hand MCP joint features are not only beneficial for 3D wrist rotations, but also for 3D elbow rotations, as 3D elbow rotations in the roll-axis are highly related to the hand MCP joints.
This is the reason why the hand MCP joint features are used for 3D rotations of all body joints, not restricted to 3D wrist rotations.

\noindent\textbf{SMPL-X shape/camera parameters.}
The shape parameter $\beta_b \in \mathbb{R}^{10}$ and 3D global translation vector $\mathbf{t}_b \in \mathbb{R}^3$ are predicted from the image feature $\mathbf{F}_b$ using the GAP and a single fully-connected layer.

\noindent\textbf{Hands and face boxes.}
The BodyNet predicts hand and face bounding boxes by concatenating the image feature $\mathbf{F}_b$ and 3D heatmap $\mathbf{H}_b$ and passing it to two convolutional layers.
The 3D heatmap $\mathbf{H}_b$ is predicted for the 3D positional pose of the body joints.
The box centers are obtained by applying the soft-argmax~\cite{sun2018integral} to the output of the two convolutional layers.
The widths and heights of the boxes are computed by performing bilinear interpolation to $\mathbf{F}_b$ at the box centers and passing the features of each box center to two fully-connected layers.

\subsection{HandNet}~\label{sec:handnet}
Like the BodyNet, the HandNet outputs 3D finger rotations $\theta_h \in \mathbb{R}^{15 \times 3}$ using Pose2Pose.
HandNet takes a hand image $\mathbf{I}_h$, cropped and resized from the high-resolution human image $\mathbf{I}$ by applying RoIAlign~\cite{he2017mask} at the predicted hand bounding box area.
Taking the hand images from the high-resolution image $\mathbf{I}$ instead of the downsampled human image $\mathbf{I}_b$ allows HandNet to utilize detailed finger information.
The hand-cropped images of the left hands are flipped to the right hands before being fed to the HandNet.
After predicting the 3D finger rotations, we flip back the outputs of the flipped left hands.
We denote the 3D finger rotations of the left and right hands as $\theta_{rh}$ and $\theta_{lh}$, respectively.

\noindent\textbf{MCP joints features to the BodyNet.}
As described in Section~\ref{sec:bodynet}, we pass $\mathbf{v}_m \in \mathbb{R}^{3*8+512*8}$, a concatenation of the 3D coordinates and features of eight MCP joints (\textit{i.e.}, four from the right hand and four from the left hand) to the BodyNet for the accurate 3D wrist rotation prediction.
To this end, we take the joint features and 3D coordinates of MCP joints of the left and right hands from the outputs of Pose2Pose.
The taken 3D coordinates and joint features are concatenated and flattened to a vector $\mathbf{v}_m \in \mathbb{R}^{3*8+512*8}$, which is passed to the BodyNet.
Among hand joints, we choose the MCP joints as they show low positional errors than other hand joints while providing essential 3D wrist rotation information.

\subsection{FaceNet}
We design FaceNet as a simple GAP-based regressor instead of Pose2Pose architecture as deformations of human faces (\textit{e.g.}, facial expressions) are not fully modeled by the 3D joint rotations.
The FaceNet predicts 3D jaw rotation $\theta_f \in \mathbb{R}^3$ and facial expression code $\psi \in \mathbb{R}^{10}$ from a face-cropped image $\mathbf{I}_f \in \mathbb{R}^{3 \times 192 \times 192}$.
The face-cropped image is cropped and resized from the high-resolution human image $\mathbf{I}$ by applying RoIAlign~\cite{he2017mask} at the predicted face bounding box area to utilize detailed face information.
We use ResNet after removing the GAP and fully-connected layer of the last part of the original ResNet.

\subsection{Loss functions}
Our framework is trained in an end-to-end manner by minimizing the loss function $L$, defined as follows.
\begin{equation}
L = L_\text{param} + L_\text{coord} + L_\text{box},
\end{equation}
where $L_\text{param}$ is a $L1$ distance between predicted and GT SMPL-X parameters.
$L_\text{coord}$ is a $L1$ distance between predicted and GT joint coordinates, and three types of joint coordinates are used to calculate the loss function: 1) 3D joint coordinates from the Pose2Pose of body and hands, 2) 3D joint coordinates, obtained by multiplying a joint regression matrix of SMPL-X to the 3D mesh, and 3) 2D joint coordinates, obtained by projecting the 3D coordinates from the 3D mesh, to the 2D space using the perspective projection.
For the projection, the predicted 3D global translation vector $\mathbf{t}_b$, fixed focal length (5000,5000), and fixed principal points (\textit{i.e.}, a center point of $\mathbf{I}_b$) are used, following ~\cite{kolotouros2019learning}.
Finally, $L_\text{box}$ is a $L1$ distance between the predicted and GT center and scale of the hands' and face's boxes.

\section{Implementation details}

PyTorch~\cite{paszke2017automatic} is used for implementation. 
The ResNet of the body branch is initialized with that of Xiao~\etal~\cite{xiao2018simple}, pre-trained on MSCOCO 2D human body pose dataset.
For the training, we use Adam optimizer~\cite{kingma2014adam} with a mini-batch size of 96.
Data augmentations, including scaling, rotation, random horizontal flip, and color jittering, are performed during the training.
All the 3D rotations are initially predicted in the 6D rotational representation of ~\cite{zhou2019continuity} and converted to the 3D axis-angle rotations.
The initial learning rate is set to $10^{-4}$ and reduced by a factor of 10 at the \nth{10} epoch.
A neutral gender SMPL-X model is used for the training and testing.
All other details will be available in our codes.
\section{Experiment}

\subsection{Datasets and evaluation metrics}
\noindent\textbf{Datasets.}
For the training, Human3.6M~\cite{ionescu2014human3}, the whole-body version of MSCOCO~\cite{jin2020whole}, MPII~\cite{andriluka20142d}, and FreiHAND~\cite{Freihand2019} are used.
The 3D pseudo-GTs for the training are obtained by NeuralAnnot~\cite{moon2022neuralannot}.
For the 3D body-only, 3D hand-only, and 3D face-only evaluations, we use 3DPW~\cite{von2018recovering}, FreiHAND~\cite{Freihand2019}, and Stirling~\cite{feng2018evaluation}, respectively.
For the 3D whole-body evaluation, we use EHF~\cite{pavlakos2019expressive} and AGORA~\cite{Patel:CVPR:2021}.
We provide qualitative results on the MSCOCO validation set.

\noindent\textbf{Evaluation metrics.}
Mean per joint position error (MPJPE) and mean per-vertex position error (MPVPE) are used to evaluate 3D joint and mesh vertices positions, respectively.
Each calculates the average 3D joint distance ($mm$) and 3D mesh vertex distance ($mm$) between the predicted and GT, after aligning a root joint translation.
The pelvis is used as the root joint when calculating the 3D errors of the whole body and body.
On the other hand, the wrists and neck are used as the root joints when calculating the 3D errors of the hands and face.
PA MPJPE and PA MPVPE further align a rotation and scale.
The 3D joint coordinates for the MPJPE and PA MPJPE are obtained by multiplying a joint regression matrix, defined in SMPL-X, to the mesh, following previous works~\cite{choutas2020monocular}.
We report average 3D errors of the left and right hands for the 3D hands' error.

\subsection{Ablation study}

\begin{figure*}[t]
\begin{center}
\includegraphics[width=\linewidth]{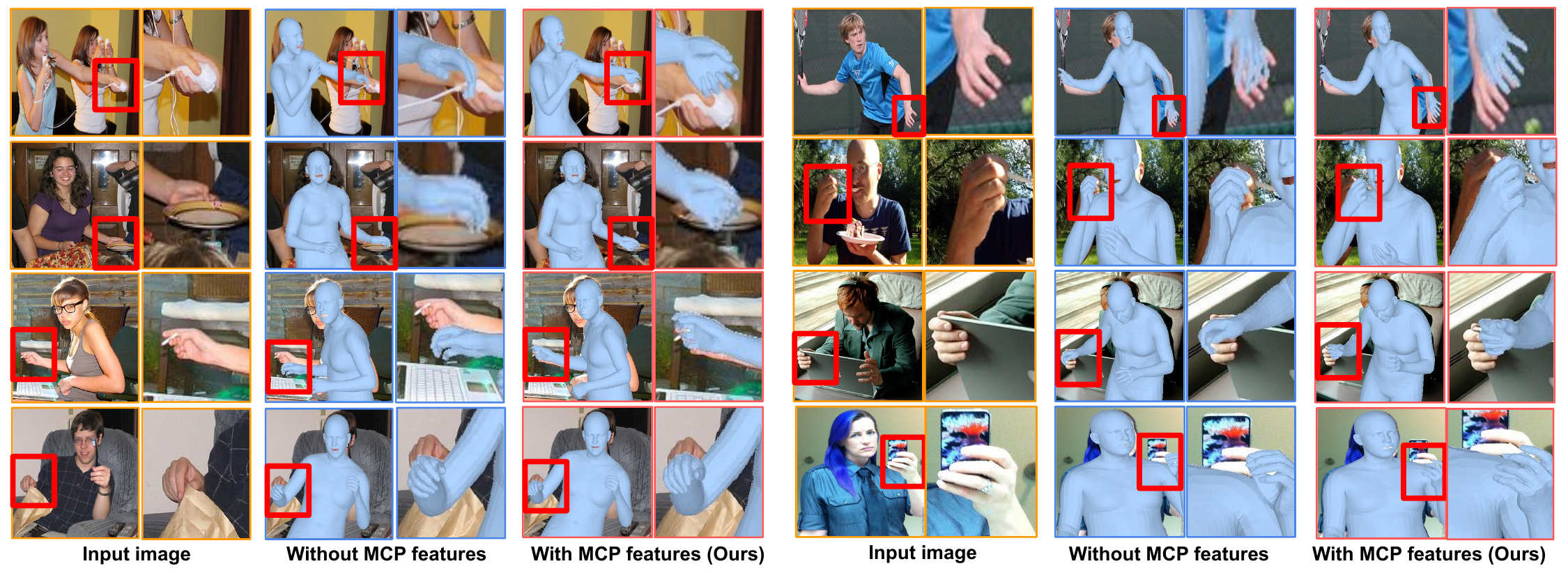}
\end{center}
\vspace*{-5mm}
   \caption{Qualitative comparison of models that doese not take and take hand MCP joint features for the 3D wrist rotation.
   Utilizing MCP features improves the 3D wrist rotation significantly.}
\vspace*{-3mm}
\label{fig:ablation_mcp}
\end{figure*}

\noindent\textbf{MCP joint features for 3D wrist rotations.}
Table~\ref{table:ablation_mcp} and Figure~\ref{fig:ablation_mcp} show that Hand4Whole produces accurate 3D wrist rotations by taking a combination of body and hand MCP joint features.
The hand MPVPE is measured by separating 3D hand meshes from a whole-body mesh without rotation alignment; therefore, wrong 3D wrist rotation significantly increases the MPVPE of hands.
To analyze the effectiveness of the hand MCP joint features, we design three variants.
They have exactly the same network architectures as ours except the BodyNet that predicts 3D body joint rotations including 3D wrist rotations.
In their BodyNet, they take the same body feature (\textit{i.e.}, a combination of 3D joint coordinates and joint features of the body) and different hand features.
Please note that the body feature is essential for plausible 3D wrist rotations when hands are occluded, as shown in Figure~\ref{fig:intro_compare}.
As they have the same HandNet architecture, all of them have similar 3D finger rotation predictions.
Therefore, most of the hand errors are from wrist rotations, not from finger rotations.

\begin{table}[t]
\small
\centering
\setlength\tabcolsep{1.0pt}
\def\arraystretch{1.1}
\begin{tabular}{C{4.8cm}|C{2.5cm}}
\specialrule{.1em}{.05em}{.05em}
Inputs for 3D wrist prediction & MPVPE (Hands) \\ \hline
Body &  50.4  \\ 
Body + Hand GAP &  43.1  \\
Body + All hand joints & 43.4 \\
\textbf{Body + MCP joints (Ours)} & \textbf{39.8} \\ \specialrule{.1em}{.05em}{.05em}
\end{tabular}
\vspace*{-3mm}
\caption{
Comparison of hands MPVPEs, obtained by models that predict 3D wrist rotation from various features, on EHF.
}
\vspace*{-3mm}
\label{table:ablation_mcp}
\end{table}

\begin{table}[t]
\small
\centering
\setlength\tabcolsep{1.0pt}
\def\arraystretch{1.1}
\begin{tabular}{C{4.5cm}|C{2.8cm}}
\specialrule{.1em}{.05em}{.05em}
Settings & PA MPVPE (Hands) \\ \hline
With body features & 12.3 \\ 
\textbf{Without body features (Ours)} & \textbf{10.8} \\ \specialrule{.1em}{.05em}{.05em}
\end{tabular}
\vspace*{-3mm}
\caption{
Comparison of hands PA MPVPEs, obtained by models that predict 3D finger rotations with and without the coarse hand information, on EHF.
}
\label{table:ablation_finger}
\vspace*{-3mm}
\end{table}

The first variant uses only the body features for the 3D wrist rotation like Zhou~\etal~\cite{zhou2021monocular}.
It suffers from inaccurate 3D wrist rotation, which indicates additional hand feature is necessary to determine 3D wrist rotation accurately.
The second variant uses a combination of the body feature and hand global average pooled (GAPed) feature for 3D wrist rotation prediction like PIXIE~\cite{feng2021collaborative}.
The hand GAP feature is obtained by performing GAP at the output of HandNet's ResNet.
It still produces less accurate 3D wrist rotation than ours, which indicates that the hand GAP feature fails to capture essential information of the 3D wrist rotations.
The third variant uses a combination of the body features and 3D joint coordinates and joint features of all two hand joints for 3D wrist rotation prediction.
It still achieves worse results than ours because the joint features of all two hand joints (\textit{i.e.}, 40 finger joints) contain too much unnecessary information considering that the body consists of 25 joints.
From an anatomical point of view, eight MCP joints of two hands mainly contribute to the 3D wrist rotation, as shown in Figure~\ref{fig:motivation}, while other finger joints often move independently with the 3D body joint rotation.
Our Hand4Whole achieves the lowest 3D errors and accurate 3D wrist rotation by taking body and MCP joint features.

\begin{table}[t]
\small
\centering
\setlength\tabcolsep{1.0pt}
\def\arraystretch{1.1}
\begin{tabular}{C{5.5cm}|C{2.5cm}}
\specialrule{.1em}{.05em}{.05em}
Inputs for 3D joint rotations & PA MPVPE (All) \\ \hline
GAP feat. & 54.8 \\ 
Joint feat. & 52.2  \\ 
2D joint coord. & 55.8  \\
3D joint coord. & 54.3 \\ 
\textbf{3D joint coord. + joint feat. (Ours)} & \textbf{50.3}  \\ \specialrule{.1em}{.05em}{.05em}
\end{tabular}
\vspace*{-3mm}
\caption{
Comparison of whole-body PA MPVPEs, obtained by models that take various input combinations for the 3D joint rotation prediction, on EHF.
}
\vspace*{-3mm}
\label{table:ablation_feature}
\end{table}

\noindent\textbf{Removing body features for 3D finger rotations.}
Table~\ref{table:ablation_finger} shows that removing the body feature produces better 3D finger rotations.
The setting that uses the body feature is similar to previous works~\cite{zhou2021monocular,feng2021collaborative}.
As PA MPVPE is calculated after the rotation alignment, aligned outputs have almost correct 3D wrist rotations.
Therefore, most of the 3D errors come from 3D fingers.
To analyze how the body feature affects the 3D finger rotation, we design a variant whose HandNet takes additional coarse hand information, obtainable from hand areas at the body feature.
To extract the coarse hand information, we perform RoIAlign~\cite{he2017mask} to the output of the first residual block in the BodyNet's ResNet using the predicted hand bounding box.
Then, the output of RoIAlign is added in an element-wise manner to the first residual block in the HandNet's ResNet.
The reason why only removing the body feature produces better 3D finger rotation is that the coarse hand information in the body feature contains various unnecessary information, such as body and backgrounds, while barely having finger information due to the small resolution of hands.
As such unnecessary information corrupts the hand feature of the HandNet, using only the hand feature from the HandNet like 3D hand-only reconstruction methods~\cite{boukhayma20193d,kulon2020weakly,zhou2020monocular}, produces better 3D finger rotations.

\begin{table}[t]
\small
\centering
\setlength\tabcolsep{1.0pt}
\def\arraystretch{1.1}
\begin{tabular}{C{3.0cm}|C{0.6cm}C{0.9cm}C{0.6cm}|C{0.8cm}C{0.9cm}C{0.6cm}}
\specialrule{.1em}{.05em}{.05em}
\multirow{ 2}{*}{Methods}  & \multicolumn{3}{c|}{PA MPVPE} & \multicolumn{3}{c}{MPVPE} \\ 
 & All & Hands & Face & All & Hands & Face \\ \hline
ExPose~\cite{choutas2020monocular} &  54.5 & 12.8 & 5.8 & 77.1 & 51.6 & 35.0 \\
FrankMocap~\cite{rong2021frankmocap} & 57.5 & 12.6 & - & 107.6 & 42.8 & - \\
PIXIE~\cite{feng2021collaborative} & 55.0 & 11.1 & \textbf{4.6} & 89.2 & 42.8 & 32.7 \\
\textbf{Hand4Whole (Ours)} & \textbf{50.3} & \textbf{10.8} & 5.8 & \textbf{76.8} & \textbf{39.8} & \textbf{26.1} \\
 \specialrule{.1em}{.05em}{.05em}
\end{tabular}
\vspace*{-3mm}
\caption{
3D errors comparison on EHF.
}
\label{table:sota_ehf}
\vspace*{-4mm}
\end{table}

\begin{table}[t]
\small
\centering
\setlength\tabcolsep{1.0pt}
\def\arraystretch{1.1}
\begin{tabular}{C{3.0cm}|C{0.6cm}C{0.9cm}C{0.6cm}|C{0.8cm}C{0.9cm}C{0.8cm}}
\specialrule{.1em}{.05em}{.05em}
\multirow{ 2}{*}{Methods}  & \multicolumn{3}{c|}{PA MPVPE} & \multicolumn{3}{c}{MPVPE} \\ 
 & All & Hands & Face & All & Hands & Face \\ \hline
ExPose~\cite{choutas2020monocular} &  88.0 & 12.1 & 4.8 & 219.8 & 115.4 & 103.5 \\
FrankMocap~\cite{rong2021frankmocap} & 90.6 & 11.2 & 4.9 & 218.0 & 95.2 & 105.4 \\
PIXIE~\cite{feng2021collaborative} & 82.7 & 12.8 & 5.4 & 203.0 & 89.9 & 95.4 \\
\textbf{Hand4Whole (Ours)} & \textbf{73.2} & \textbf{9.7} & \textbf{4.7} & \textbf{183.9} & \textbf{72.8} & \textbf{81.6} \\
 \specialrule{.1em}{.05em}{.05em}
\end{tabular}
\vspace*{-3mm}
\caption{
3D errors comparison on AGORA.
}
\label{table:sota_agora}
\vspace*{-4mm}
\end{table}

\noindent\textbf{Effectiveness of joint features.}
Table~\ref{table:ablation_feature} shows that the joint feature of our Pose2Pose is much more beneficial than previous GAPed feature~\cite{choutas2020monocular,rong2021frankmocap,feng2021collaborative}.
In addition, combining the joint feature with 3D joint coordinates like ours achieves the best results.
For the demonstration, we designed four variants that take different combinations of features for the 3D body and hand joint rotation prediction.
The first and second variants take a GAP feature vector and joint feature, respectively.
The GAP feature is obtained by performing GAP to the output of ResNet.
As GAP marginalizes the spatial domain, the feature vector losses detailed human articulation information.
On the other hand, our joint feature preserves such detailed articulation information by interpolating human joint positions, beneficial for accurate 3D joint rotation prediction.
The third and fourth variants take 2D and 3D joint coordinates, respectively.
Compared to the third variant, the fourth variant achieves lower PA MPVPE by utilizing additional depth information.
Finally, we combine the 3D joint coordinates and joint features for the best results.

\subsection{Comparison with state-of-the-art methods}~\label{sec:compare_sota}

\noindent\textbf{EHF (Whole-body evaluation benchmark).}
Table~\ref{table:sota_ehf} shows that our Hand4Whole largely outperforms all previous whole-body methods on EHF.
As existing works mainly have reported PA metrics, we use their released codes and pre-trained weights to report MPVPEs.

\noindent\textbf{AGORA (Whole-body evaluation benchmark).}
Table~\ref{table:sota_agora} shows that our Hand4Whole largely outperforms all previous whole-body methods on AGORA.
We obtained all numbers using their released codes with pre-trained weights and an official evaluation tool\footnote{\url{https://github.com/pixelite1201/agora_evaluation}}.
We use the same human bounding boxes for all methods.

\begin{table}[t]
\small
\centering
\setlength\tabcolsep{1.0pt}
\def\arraystretch{1.1}
\begin{tabular}{C{4.5cm}C{1.2cm}C{1.7cm}}
\specialrule{.1em}{.05em}{.05em}
Methods & MPJPE & PA MPJPE \\ \hline
\multicolumn{3}{l}{\textbf{* Body-only methods}} \\
HMR~\cite{kanazawa2018end} & 130.0 & 81.3 \\
SPIN~\cite{kolotouros2019learning} & 96.9 & 59.2 \\
Pose2Mesh~\cite{choi2020p2m} & \textbf{88.9} & 58.3 \\ 
I2L-MeshNet~\cite{moon2020i2l} & 93.2 & 57.7 \\ 
ROMP~\cite{ROMP} & 91.3 & \textbf{54.9} \\ \hline
\multicolumn{3}{l}{\textbf{* Whole-body methods}} \\
ExPose~\cite{choutas2020monocular} & 93.4 & 60.7 \\
FrankMocap~\cite{rong2021frankmocap} & - & 61.9 \\
PIXIE~\cite{feng2021collaborative} & - & 61.3 \\
\textbf{Hand4Whole (Ours)} & \textbf{86.6} & \textbf{54.4} \\
 \specialrule{.1em}{.05em}{.05em}
\end{tabular}
\vspace*{-3mm}
\caption{3D body error comparison on 3DPW.}
\label{table:sota_3dpw}
\vspace*{-4mm}
\end{table}

\begin{table}[t]
\small
\centering
\setlength\tabcolsep{1.0pt}
\def\arraystretch{1.1}
\begin{tabular}{C{3.7cm}C{2.0cm}C{2.0cm}}
\specialrule{.1em}{.05em}{.05em}
 Methods & PA errors &  F scores \\ \hline
 \multicolumn{3}{l}{\textbf{* Hand-only methods}} \\
 FreiHAND~\cite{Freihand2019} & 10.7 / - & 0.529 / 0.935 \\
 Pose2Mesh~\cite{choi2020p2m} & 7.8 / 7.7 & 0.674 / 0.969 \\
 I2L-MeshNet~\cite{moon2020i2l}  & 7.6 / 7.4 & 0.681 / 0.973 \\ 
 METRO~\cite{lin2021end} & \textbf{6.7} / \textbf{6.8} & \textbf{0.717} / \textbf{0.981} \\ \hline
 \multicolumn{3}{l}{\textbf{* Whole-body methods}} \\
 ExPose~\cite{choutas2020monocular} & 11.8 / 12.2 & 0.484 / 0.918 \\
 Zhou~\etal~\cite{zhou2021monocular} & - / 15.7 & - / - \\
 FrankMocap~\cite{rong2021frankmocap} & 11.6 / 9.2 & 0.553 / 0.951 \\
 PIXIE~\cite{feng2021collaborative} & 12.1 / 12.0 & 0.468 / 0.919 \\
 \textbf{Hand4Whole (ResNet-18)} & 8.6 / 8.6  & 0.621 / 0.962  \\
 \textbf{Hand4Whole (Ours)} & \textbf{7.7} / \textbf{7.7} & \textbf{0.664} / \textbf{0.971}  \\
 \specialrule{.1em}{.05em}{.05em}
\end{tabular}
\vspace*{-3mm}
\caption{
3D hand errors (PA MPVPE/PA MPJPE and F-score@5mm/15mm) comparison on FreiHAND.
}
\label{table:sota_freihand}
\vspace*{-4mm}
\end{table}

\begin{figure*}[t]
\begin{center}
\includegraphics[width=0.9\linewidth]{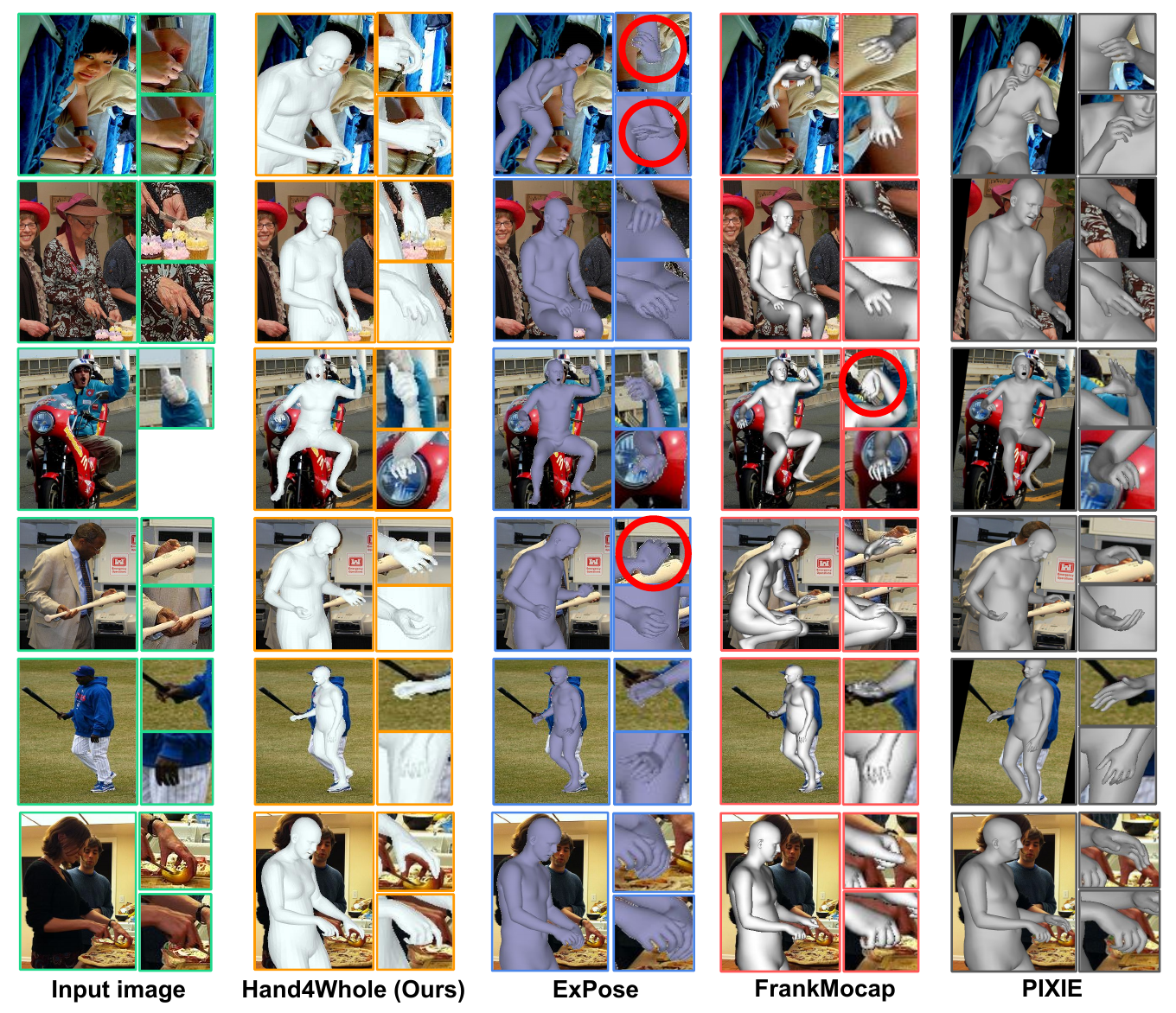}
\end{center}
\vspace*{-7mm}
   \caption{Qualitative comparison of the proposed Hand4Whole, ExPose~\cite{choutas2020monocular}, FrankMocap~\cite{rong2021frankmocap}, and PIXIE~\cite{feng2021collaborative}.
   Implausible 3D wrist rotations are highlighted.}
\vspace*{-3mm}
\label{fig:qualitative_comparison}
\end{figure*}

\noindent\textbf{3DPW (Body-only evaluation benchmark).}
Table~\ref{table:sota_3dpw} shows that our Hand4Whole largely outperforms all previous whole-body methods on 3DPW, although it is slightly beaten by recent body-only methods.
The stronger performance on the 3D body benchmark than previous whole-body methods results from a combination of 3D joint coordinates and joint features for the 3D body joint rotation prediction.
On the other hand, all previous whole-body methods use GAP features for the 3D body joint rotation prediction.

\noindent\textbf{FreiHAND (Hand-only evaluation benchmark).}
Table~\ref{table:sota_freihand} shows that our Hand4Whole largely outperforms all previous whole-body methods on FreiHAND, although it is slightly beaten by recent hand-only methods.
We obtained the results of FrankMocap using their released codes and pre-trained weights as they did not report their results on FreiHAND.
In particular, Hand4Whole achieves the best results even after changing our hand branch ResNet-50 to ResNet-18 following ExPose~\cite{choutas2020monocular}.

\noindent\textbf{MSCOCO (Qualitative comparison of the whole-body).}
Figure~\ref{fig:qualitative_comparison} shows that our Hand4Whole produces much more accurate 3D wrists and finger rotations than previous whole-body methods.

Overall, Hand4Whole largely outperforms all previous whole-body methods.
In particular, it produces much better 3D hand results.
The comparisons are consistent with the results of the ablation studies.

\section{Conclusion}
We present Hand4Whole, a whole-body 3D human mesh estimation system that produces much better 3D hands in the whole-body 3D mesh.
Hand4Whole utilizes hand MCP joint features for the 3D wrist rotation by introducing Pose2Pose.
In addition, it discards the body feature when predicting 3D finger rotations.
Hand4Whole largely outperforms previous 3D whole-body human mesh estimation methods on all benchmarks.

\noindent\textbf{Acknowledgements.}
This work was supported in part by IITP grant funded by the Korea government (MSIT) [No. 2021-0-01343, Artificial Intelligence Graduate School Program (Seoul National University)]

\clearpage

\begin{center}
\textbf{\large Supplementary Material for\\``Accurate 3D Hand Pose Estimation for \\Whole-Body 3D Human Mesh Estimation"}
\end{center}

\setcounter{figure}{0}
\setcounter{table}{0}
\setcounter{section}{0}

\renewcommand{\thefigure}{\Alph{figure}}
\renewcommand{\thetable}{\Alph{table}}
\renewcommand{\thesection}{\Alph{section}}

In this supplementary material, we present more experimental results that could not be included in the main manuscript due to the lack of space.

\section{Qualitative comparisons}

\subsection{MSCOCO}
\begin{figure*}[t]
\begin{center}
\includegraphics[width=\linewidth]{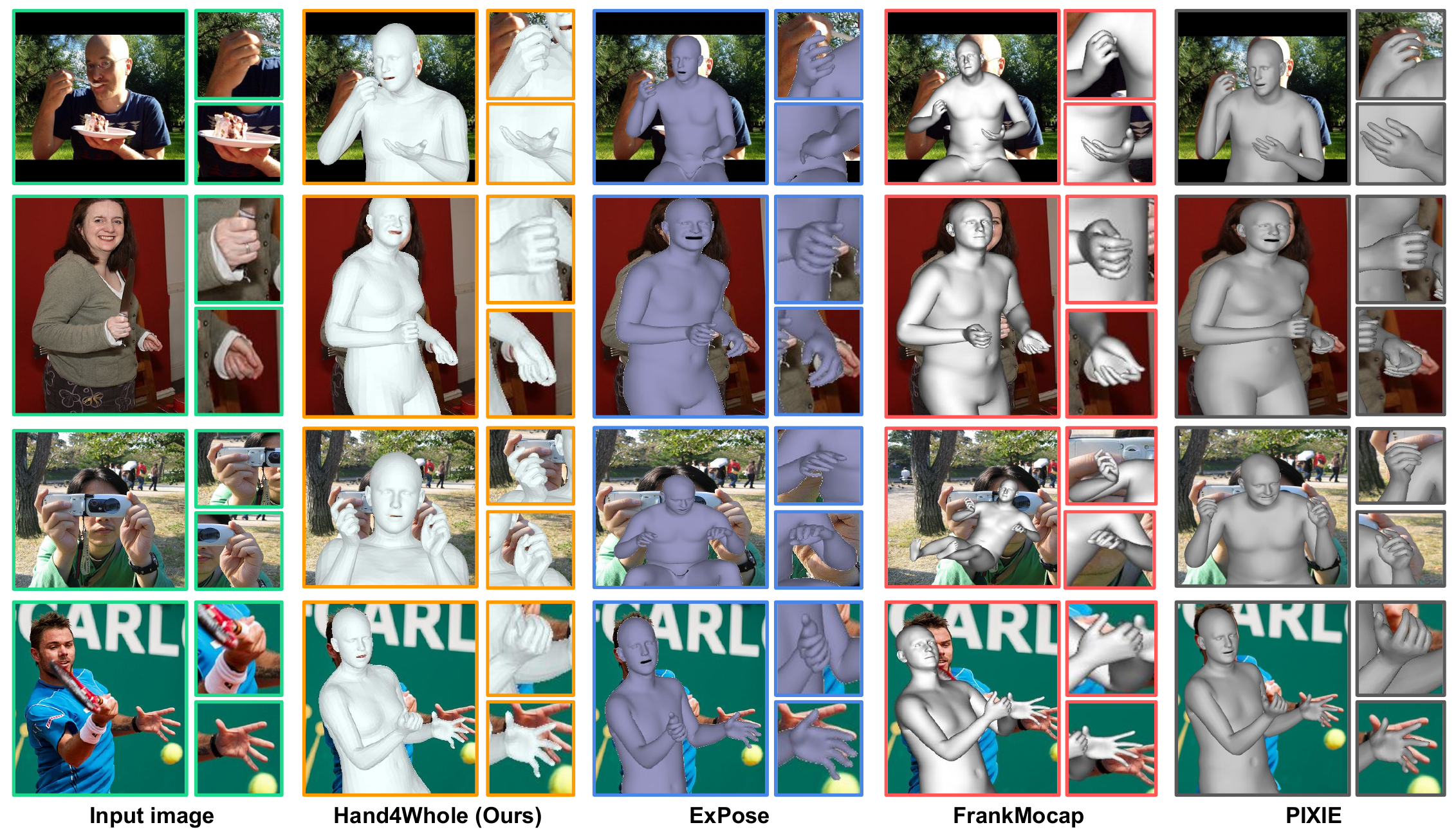}
\end{center}
\vspace*{-5mm}
   \caption{Qualitative comparison of the proposed Hand4Whole, ExPose~~\cite{choutas2020monocular}, FrankMocap~\cite{rong2021frankmocap}, and PIXIE~\cite{feng2021collaborative} on MSCOCO.}
\vspace*{-3mm}
\label{fig:qualitative_comparison_1}
\end{figure*}

\begin{figure*}[t]
\begin{center}
\includegraphics[width=\linewidth]{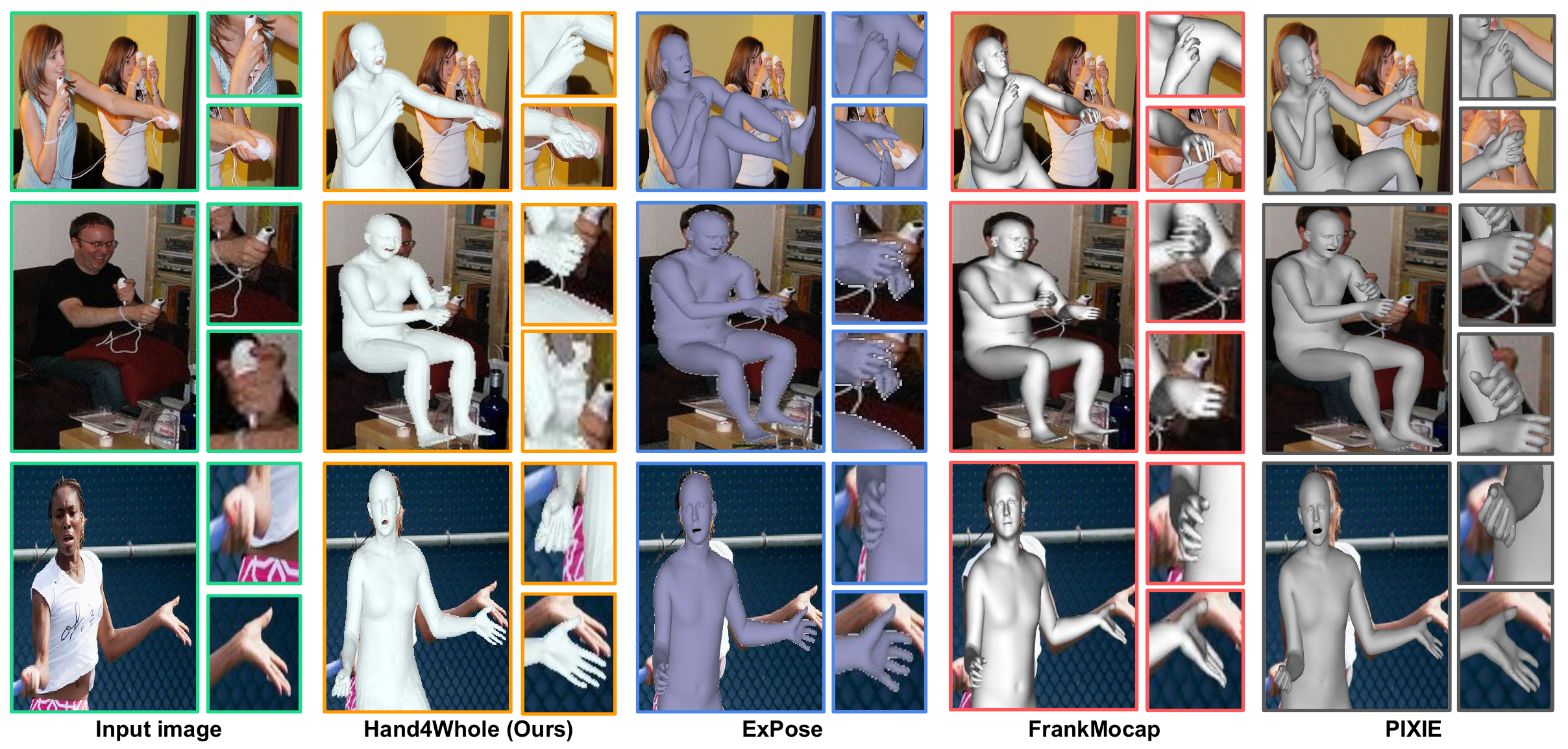}
\end{center}
\vspace*{-5mm}
   \caption{Qualitative comparison of the proposed Hand4Whole, ExPose~~\cite{choutas2020monocular}, FrankMocap~\cite{rong2021frankmocap}, and PIXIE~\cite{feng2021collaborative} on MSCOCO.}
\vspace*{-3mm}
\label{fig:qualitative_comparison_2}
\end{figure*}

Figure~\ref{fig:qualitative_comparison_1} and ~\ref{fig:qualitative_comparison_2} show that our Hand4Whole produces more accurate results on in-the-wild images of MSCOCO.
In particular, ours produce much better 3D hands results.

\subsection{3DPW}
The video in this link\footnote{\url{https://www.youtube.com/watch?v=Ym_CH8yxBso}} shows that our Hand4Whole produces more accurate and plausible expressive 3D human mesh than ExPose~\cite{choutas2020monocular} and FrankMocap~\cite{rong2021frankmocap} on videos of 3DPW~\cite{von2018recovering}.
In particular, ours achieves much better and stable hands results when hands are invisible by using body and hand MCP joint features.
On the other hand, ExPose and FrankMocap do not use body features, which results in implausible 3D hands.
Hand4Whole, ExPose, and FrankMocap are run on every single frame without leveraging temporal information.
We did not apply any post-processing, such as average filtering, on the outputs.
The results of ExPose and FrankMocap are obtained by their officially released codes.

\subsection*{License of the Used Assets}

\begin{compactitem}[$\bullet$]
    \item MSCOCO dataset~\cite{lin2014microsoft} belongs to the COCO Consortium and are licensed under a Creative Commons Attribution 4.0 License.
    \item Human3.6M dataset~\cite{ionescu2014human3}'s licenses are limited to academic use only. 
    \item MPII dataset~\cite{andriluka20142d} is released for academic research only and it is free to researchers from educational or research institutes for non-commercial purposes.
    \item 3DPW dataset~\cite{von2018recovering} is released for academic research only and it is free to researchers from educational or research institutes for non-commercial purposes.
    \item FreiHAND dataset~\cite{Freihand2019} is released for academic research only and it is free to researchers from educational or research institutes for non-commercial purposes.
    \item FFHQ dataset~\cite{karras2019style}'s individual images were published in Flickr by their respective authors under either Creative Commons BY 2.0, Creative Commons BY-NC 2.0, Public Domain Mark 1.0, Public Domain CC0 1.0, or U.S. Government Works license.
    The dataset itself (including JSON metadata, download script, and documentation) is made available under Creative Commons BY-NC-SA 4.0 license by NVIDIA Corporation.
    \item Stirling dataset~\cite{feng2018evaluation} is released for academic research only and it is free to researchers from educational or research institutes for non-commercial purposes.
    \item EHF dataset~\cite{pavlakos2019expressive} is released for academic research only and it is free to researchers from educational or research institutes for non-commercial purposes.
    \item AGORA dataset~\cite{Patel:CVPR:2021} is released for academic research only and it is free to researchers from educational or research institutes for non-commercial purposes.
    \item \href{https://github.com/vchoutas/expose}{ExPose~\cite{choutas2020monocular} codes} are released for academic research only and it is free to researchers from educational or research institutes for non-commercial purposes.
    \item \href{https://github.com/facebookresearch/frankmocap}{FrankMocap~\cite{rong2021frankmocap} codes} are CC-BY-NC 4.0 licensed.
    \item \href{https://github.com/YadiraF/PIXIE}{PIXIE~\cite{feng2021collaborative} codes} are released for academic research only and it is free to researchers from educational or research institutes for non-commercial purposes.
\end{compactitem}

\clearpage
\clearpage

{\small
\bibliographystyle{ieee_fullname}
\bibliography{main}
}

\end{document}